\documentclass{article}

\usepackage[final,main]{neurips_2026}

\usepackage[utf8]{inputenc}   
\usepackage[T1]{fontenc}      
\usepackage{hyperref}         
\usepackage{url}              
\usepackage{booktabs}         
\usepackage{amsfonts}         
\usepackage{amsmath}          
\usepackage{amssymb}          
\usepackage{nicefrac}         
\usepackage{microtype}        
\usepackage{xcolor}           
\usepackage{graphicx}         
\usepackage{tikz}             

\title{Learning to Rotate: Temporal and Semantic Rotary Encoding for Sequential Modeling}

\author{%
  Hailing Cheng \\
  LinkedIn Inc.\\
  Mountain View, CA 94043 \\
  \texttt{haicheng@linkedin.com} \\
  \And
  Daqi Sun \\
  LinkedIn Inc.\\
  Mountain View, CA 94043 \\
  \texttt{daqsun@linkedin.com} \\
  \And
  Xinyu Lu \\
  LinkedIn Inc.\\
  Mountain View, CA 94043 \\
  \texttt{xinlu@linkedin.com} \\
}

\begin{document}

\maketitle

\begin{abstract}
Every Transformer architecture dedicates enormous capacity to learning rich representations in \emph{semantic embedding space}---yet the rotation manifold acted upon by Rotary Positional Embeddings (RoPE) has been treated as a fixed, hand-crafted structure, populated only by discrete ordinal indices. We argue that this \emph{rotation space} is a largely overlooked second dimension of expressivity in the attention mechanism, one whose systematic exploration may open a new door for attention-based architectures. The analogy to complex numbers is instructive: just as introducing the imaginary axis---orthogonal to and independent of the real line---unlocked new algebraic structure once believed impossible, treating the rotation manifold as a learnable, signal-conditioned space opens an orthogonal degree of freedom in attention. In this framing, the token embedding encodes the \emph{semantic} (real) component of a representation---\emph{what} a token means---while the rotation encodes its \emph{dynamic} (imaginary) component---\emph{how} it relates to every other token across time, position, and context.

We introduce \textbf{SIREN-RoPE}, a concrete instantiation of this idea, which populates the rotation dimension with heterogeneous signals---continuous timestamps, cyclical temporal patterns, and categorical metadata---via a dual-branch Sinusoidal Representation Network (SIREN) \citep{sitzmann2020siren}. As a proof of concept, we evaluate on a production-scale news feed dataset from a major social network using a generative recommender as the ranking model, demonstrating that activating this hidden dimension yields consistent improvements across calibration and ranking objectives with negligible computational overhead. We invite the community to view the rotation space not as a solved positional-encoding detail, but as an untapped axis whose rich structure may prove as consequential for attention as the imaginary unit proved for algebra.
Demo code is available at \url{https://github.com/hailingc/siren_rope}.
\end{abstract}

\section{Introduction}
\label{sec:intro}

Rotary Position Embeddings (RoPE) \citep{su2024roformer} in Large Language Models (LLMs) 
encode sequence order as planar rotations of query/key subspaces, yielding strong
relative-position modeling. Yet the rotation manifold itself has always been a
fixed, hand-crafted structure driven by discrete ordinal indices. In sequential
recommendation and event-stream modeling, this is a poor proxy: an interaction
seven days ago is qualitatively different from one seven minutes ago, even at the same ordinal offset.

We argue this points to a deeper opportunity. Just as extending the real line to the complex plane---by adding an orthogonal imaginary axis---unlocked entirely new algebraic structure, treating the RoPE rotation manifold as a learnable, signal-conditioned space opens an orthogonal expressive dimension in attention. Token embeddings encode the \emph{semantic} (real) component---\emph{what} a token means; the rotation encodes the \emph{dynamic} (imaginary) component---\emph{how} it relates to every other token across time and context. This dimension has been almost entirely unexplored.

\subsection{Contributions}

We introduce \textbf{SIREN-RoPE}, which treats the rotation manifold as a first-class learnable space:
\begin{itemize}
  \item \textbf{Temporal rotation.} A dual-branch SIREN--DNN network maps continuous timestamps into per-dimension rotation angles, capturing both periodic (daily/weekly) and aperiodic (recency decay) temporal structure directly in the geometry of attention.
  \item \textbf{Adaptive frequency learning.} Per-dimension frequency scales and an ordinal-gate scalar $\lambda$ are jointly learned, replacing hand-crafted inverse-frequency constants.
  \item \textbf{Empirical validation.} On a production-scale social feed dataset, SIREN-RoPE consistently improves calibration (NE) and ranking (AUC) across three engagement tasks with ${\sim}0.2\%$ extra parameters. Ablations confirm the rotation dimension carries signal complementary to and independent of the embedding space.
\end{itemize}

\section{Related Work}

\paragraph{Positional Encodings and RoPE.}
Early Transformers inject order via fixed sinusoidal or learned absolute embeddings \citep{vaswani2017attention}; subsequent work shifted to relative formulations for better length generalization \citep{shaw2018relative, dai2019transformerxl, raffel2020t5}. RoPE encodes position as planar rotations in query-key subspaces, expressing relative offsets as phase differences while preserving vector norms \citep{su2024roformer}. Extensions such as YaRN and LongRoPE focus on length extrapolation via interpolation \citep{peng2023yarn, zhang2024longrope}. All of these treat the rotation manifold as a fixed positional scaffold; we instead treat it as a learnable, signal-conditioned space.

\paragraph{Temporal Dynamics in Sequential Models.}
Methods like Time2Vec \citep{kazemi2019time2vec} and TiSASRec \citep{li2020tisasrec} incorporate pairwise time intervals and periodic embeddings to capture non-linear temporal dynamics. DeBERTa \citep{he2020deberta} and TUPE \citep{ke2020tupe} disentangle content from positional context in language modeling. Most closely related to our work, TO-RoPE \citep{shirkavand2025torope} applies the same inverse-frequency schedule as standard RoPE but substitutes the raw timestamp $T_i$ for the ordinal index, yielding a per-dimension rotation angle $T_i \cdot \theta_j$. SIREN-RoPE generalizes this idea by replacing the fixed inverse-frequency mapping with a dual-branch SIREN--DNN network that maps rich multi-scale temporal features into learned rotation angles, enabling the model to capture complex periodicities that the fixed frequency schedule cannot express.

\paragraph{Implicit Neural Representations and Continuous-Time Models.}
SIRENs \citep{sitzmann2020siren} use periodic activations with principled initialization to overcome spectral bias \citep{tancik2020fourier}, enabling multi-scale periodic function learning. INR-based methods have been adapted for time-series \citep{fons2022hypertime, lenaour2024inrts} but not for attention geometry. Neural ODEs/CDEs \citep{chen2018neuralode, kidger2020neuralcde} and hybrid models like ContiFormer \citep{chen2024contiformer} handle continuous-time sequences at significant inference cost. SIREN-RoPE injects continuous-time inductive bias directly into the rotary angle, achieving comparable expressivity at the cost of a small MLP.

\section{SIREN-RoPE: Unified Rotary Encoding}
\label{sec:method}

\subsection{Problem Formulation}

Let $\mathcal{S} = \{(\mathbf{e}_i, T_i)\}_{i=1}^{C}$ denote a user interaction sequence of
length $C$, where $\mathbf{e}_i \in \mathbb{R}^{d}$ is the item embedding and $T_i \in \mathbb{R}$ represents
the interaction timestamp. Conventional Rotary Positional Embedding (RoPE) \citep{su2024roformer} modulates
the inner product between query and key representations using rotations derived from discrete ordinal indices $p_i$.
While the attention weights are primarily driven by the semantic content of $\mathbf{e}_i$, this formulation constrains
the relative positional dependency to be a strictly periodic function of the ordinal displacement $|p_i - p_j|$.
Consequently, it neglects the irregular temporal intervals $\Delta T = T_i - T_j$ that often characterize real-world user behavior.

In social-feed recommendation, treating ordinal position as a proxy for time is fundamentally misaligned with empirical user behavior.
An interaction at 08:00 on a Monday is qualitatively distinct from one at 22:00 on a Saturday, even when they share the same ordinal
displacement $|p_i - p_j|$. Human behavior is governed by strong periodicities---\textbf{24-hour diurnal cycles} and \textbf{7-day weekly
rhythms}---so user intent and content relevance are tied to absolute temporal coordinates, a structure that ordinal indices alone cannot capture.
We therefore seek a rotation angle
$\Theta_j(T_i, p_i)$ satisfying three criteria:
\begin{enumerate}
  \item \textbf{Temporal richness:} captures multi-scale cyclical patterns
    from $T_i$.
  \item \textbf{Ordinal preservation:} retains the monotone recency-decay and
    translational equivariance properties of standard RoPE.
  \item \textbf{Adaptive balance:} allows end-to-end gradient descent to
    determine how much each component contributes.
\end{enumerate}

\subsection{Ordinal-Temporal Fusion}

SIREN-RoPE replaces the fixed angle $p_i \theta_j$ with:
\begin{equation}
  \boxed{
    \Theta_j(T_i,\, p_i)
    \;=\;
    \underbrace{f_{\phi}(T_i)_j \cdot \omega^{s}_j}_{\text{Temporal (SIREN)}}
    \;+\;
    \underbrace{p_i \cdot \theta_j \cdot \lambda}_{\text{Ordinal (scaled)}}
  }
  \label{eq:siren_rope}
\end{equation}
where:
\begin{itemize}
  \item $i \in \{0,\ldots,C-1\}$ is the item's ordinal index in the sequence;
  \item $p_i = i$ is the ordinal position of item $i$;
  \item $f_{\phi}: \mathbb{R}^{d_t} \to \mathbb{R}^{d_k/2}$ is the
    dual-branch SIREN network (Section~\ref{sec:siren_arch}), mapping
    $d_t$-dimensional timestamp features to rotation angles;
  \item $\omega^{s}_j \in \mathbb{R}^{d_k/2}$ are learnable per-dimension
    frequency scalings;
  \item $\theta_j = \mathrm{base}^{-2j/d_k}$ are the fixed inverse frequencies
    used in standard RoPE;
  \item $\lambda \in \mathbb{R}$ is a learnable scalar gate (initialized
    to $1.0$) controlling the ordinal contribution.
\end{itemize}

The rotation is then applied identically to standard RoPE with $\Theta_j$
substituted for $p\theta_j$, where $\mathbf{q}$ is the query vector and
$\mathbf{k}$ is the key vector:
\begin{align}
  q_{2i}' &= q_{2i} \cos(\Theta_j) - q_{2i+1} \sin(\Theta_j),
  \label{eq:rot_cos}\\
  q_{2i+1}' &= q_{2i} \sin(\Theta_j) + q_{2i+1} \cos(\Theta_j).
  \label{eq:rot_sin}
\end{align}
The same rotation applies to $\mathbf{k}$. This real-number formulation avoids
complex-number operations and is fully compatible with \texttt{torch.compile}.

\paragraph{Initialization and convergence.} At initialization
($\omega^s_j = \pi$, $\lambda = 1.0$), the model begins close to a scaled
form of standard RoPE augmented by a SIREN offset, providing a stable
starting point. As training progresses, $\omega^s_j$ and $\lambda$ learn to
balance temporal and ordinal contributions. In practice,
we observe a striking pattern: without temporal signals (ordinal RoPE only),
$\lambda$ remains near its initial value of $1.0$ throughout training.
Upon introducing SIREN-RoPE, however, $\lambda$ drops to $0.044$ at convergence,
indicating that the model has learned to rely predominantly on temporal
modulation over ordinal position---the rotation dimension, once given a rich
temporal signal to carry, largely supplants the need for discrete ordinal
recency.

\subsection{Dual-Branch SIREN Architecture}
\label{sec:siren_arch}

$f_\phi$ additively combines a periodic SIREN branch and an aperiodic DNN branch:
\begin{equation}
  f_\phi(T) \;=\;
    \underbrace{f_{\mathrm{sin}}(T)}_{\text{Periodic branch}}
    \;+\;
    \underbrace{f_{\mathrm{DNN}}(T)}_{\text{Aperiodic branch}}.
  \label{eq:dual_branch}
\end{equation}

\textbf{Periodic branch $f_{\mathrm{sin}}$} uses the SIREN architecture \citep{sitzmann2020siren},
where each hidden layer computes $\sin(\omega_0 \mathbf{W}\mathbf{x} + \mathbf{b})$.
The sine activations enable the branch to
autonomously discover temporal periodicities beyond the manually specified daily/weekly cycles.

\textbf{Aperiodic branch $f_{\mathrm{DNN}}$} uses standard ReLU activations to capture monotone trends such as content-recency
decay. When the input features are themselves cyclical (as with the $(\cos,\sin)$ pairs in Eq.~\eqref{eq:temp_feat}), the DNN branch
retains the flexibility to model periodic components as well.

Full implementation details are available at \url{https://github.com/hailingc/siren_rope}.

\subsection{Temporal Input Features}

The input to $f_\phi$ is a 5-dimensional decomposition of the raw
Unix timestamp $T$:
\begin{equation}
  \mathbf{t}(T) = \Bigl[
    \cos\!\Bigl(\frac{2\pi T}{\tau_d}\Bigr),\;
    \sin\!\Bigl(\frac{2\pi T}{\tau_d}\Bigr),\;
    \cos\!\Bigl(\frac{2\pi T}{\tau_w}\Bigr),\;
    \sin\!\Bigl(\frac{2\pi T}{\tau_w}\Bigr),\;
    \tilde{T}
  \Bigr],
  \label{eq:temp_feat}
\end{equation}
where $\tau_d = 86{,}400\,\mathrm{s}$ (daily periodicity), $\tau_w =
604{,}800\,\mathrm{s}$ (weekly periodicity), and $\tilde{T}$ is a normalized
long-range offset. Representing each cycle as a $(\cos, \sin)$ pair ensures
continuity across period boundaries (midnight, end of week) and avoids
phase-discontinuity artifacts. Notably, in this rotational representation the model need not know the user's timezone to capture morning or evening patterns; the periodicity is recovered automatically from the behavioral sequence.

\section{Experiments}
\label{sec:experiments}

\subsection{Experimental Setup}

\paragraph{Dataset.}
We evaluate on a production-scale dataset collected from the news feed of a
major social network, comprising user--item interaction events from one year
of production logs. Each user's history is represented as an interleaved
sequence $[\mathbf{e}_1, \mathbf{a}_1, \mathbf{e}_2, \mathbf{a}_2, \ldots,
\mathbf{e}_C, \mathbf{a}_C]$, where $\mathbf{e}_i$ is the item representation
and $\mathbf{a}_i$ is the corresponding action representation; the action
labels serve as prediction targets. Full details of this input format are
described in \citet{cheng2026}.
Each event carries a Unix timestamp, user and item features,
and engagement labels across three tasks: \textbf{Contribution} (like, comment or
share actions), \textbf{Like}, and \textbf{LongDwell} (sustained dwell-time
engagement). Temporal patterns are pronounced due to weekday/weekend and
morning/evening behavioral cycles. The dataset reflects production-level
complexity in terms of interaction volume, item diversity, and behavioral
heterogeneity, representing conditions substantially more demanding than
standard academic benchmarks. Our goal is not to establish a new benchmark
comparison, but to demonstrate on a real production system that the rotation
dimension carries predictive signal worth the community's attention.

\paragraph{Baselines.}
All four models (three baselines and SIREN-RoPE) share the identical AttnMVP
backbone, feature set, and training procedure.  The \emph{only} controlled
variable is how---or whether---the item timestamp is used:

\begin{itemize}
  \item \textbf{Ordinal RoPE}: Standard RoPE with discrete ordinal position
    indices. \emph{No timestamp is used anywhere in the model.}
    This is the production baseline.
  \item \textbf{Timestamp-as-Feature}: Standard ordinal RoPE, but the
    5-dimensional timestamp features $\mathbf{t}(T)$
    (Eq.~\eqref{eq:temp_feat}) are appended to the item sequence features
    before the Transformer stack. The rotation angle remains purely ordinal.
    This baseline isolates whether the timestamp carries useful signal when
    routed through the \emph{embedding} (semantic) dimension rather than the
    rotation dimension.
  \item \textbf{TO-RoPE} (Time and Order RoPE) \citep{shirkavand2025torope}:
    Applies the same inverse-frequency schedule as standard RoPE
    ($\theta_j = \mathrm{base}^{-2j/d_k}$) but uses the normalized timestamp
    $\tilde{T}_i$ as the index instead of the ordinal position, yielding
    rotation angle $\tilde{T}_i \cdot \theta_j$ per dimension. No timestamp is
    added to the sequence features. This baseline tests whether adding static
    time frequencies in the standard RoPE formula suffices.
\end{itemize}

\begin{figure}
  \centering
  \includegraphics[width=0.4\columnwidth]{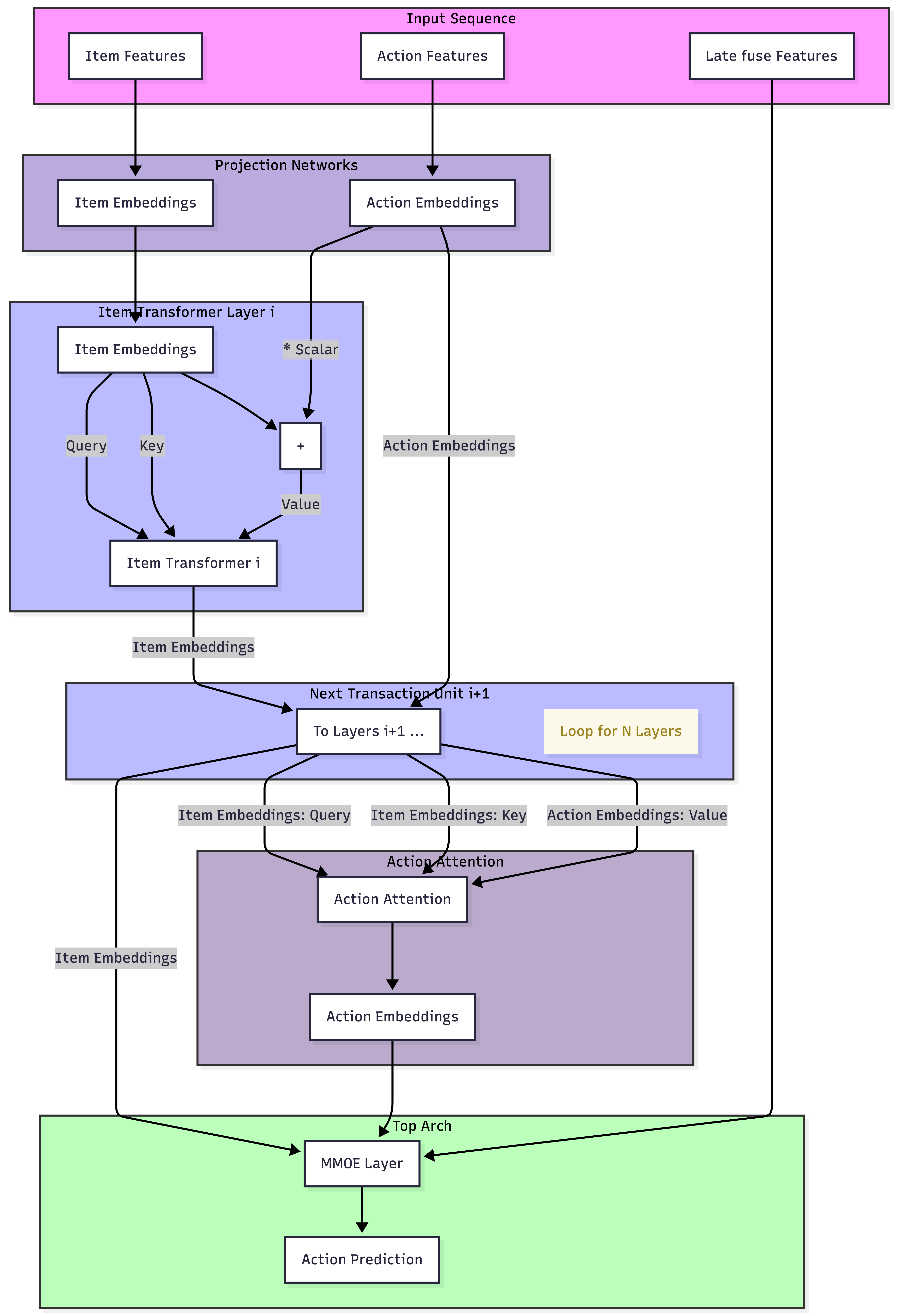}
  \caption{Base model architecture (AttnMVP, \citet{cheng2026}). Item and action embeddings
    are maintained as separate streams. Item embeddings form the Query/Key projections in
    every Transformer layer while action embeddings are mixed into the Value stream under
    strict causal masking. A final action-pooling step aggregates historical action signals
    conditioned on item-level similarity before the prediction head.}
  \label{fig:model_arch}
\end{figure}

\paragraph{Base Model Architecture.}
All experiments use the \textbf{AttnMVP} (Attention-based Mixed Value Pooling) backbone
proposed by \citet{cheng2026}; Figure~\ref{fig:model_arch} provides an overview.
AttnMVP maintains item and action embeddings as separate streams, departing from the
interleaved token formulation of prior generative recommenders \citep{zhai2024actions}.
Within each Transformer layer, item embeddings serve as both Queries and Keys, while action
embeddings are mixed additively into the value stream:
$\mathbf{V}_t = \mathbf{H}_t + \alpha\,\mathbf{a}_t$, where $\mathbf{H}_t$ is the current
item representation and $\alpha$ controls the action contribution. Strict causality is
enforced via a \emph{query-shifting} mechanism compatible with FlashAttention: queries are
shifted one step left relative to the keys, so the representation of item $i_n$ attends
only to positions $\{0,\ldots,n-1\}$, preventing target-label leakage through the value
stream. After the final Transformer layer, a causally masked action-pooling step aggregates
historical action embeddings conditioned on item similarity, producing the action-aware
vector fed to the prediction head.

Every model variant in this paper uses this identical backbone with the identical item,
action, and context feature set.  The sole controlled variable across models is how
the timestamp is incorporated---Table~\ref{tab:model_comparison} summarizes the design
space.  For SIREN-RoPE and TO-RoPE the timestamp enters \emph{only} through the
rotation angle and is never added to the sequence features.  For the
Timestamp-as-Feature baseline the timestamp enters \emph{only} through the sequence
features and the rotation angle remains standard ordinal RoPE.  Ordinal RoPE uses no
timestamp anywhere.

\begin{table}[h]
  \centering
  \small
  \caption{How each model incorporates the item timestamp. All other
    components (backbone, features, training) are identical.}
  \label{tab:model_comparison}
  \begin{tabular}{lcc}
    \toprule
    Model & Timestamp in sequence features & Timestamp in rotation angle \\
    \midrule
    Ordinal RoPE           & \texttimes & \texttimes \\
    Timestamp-as-Feature   & \checkmark & \texttimes \\
    TO-RoPE                & \texttimes & time-domain inverse frequencies \\
    SIREN-RoPE (ours)      & \texttimes & dual-branch SIREN \\
    \bottomrule
  \end{tabular}
\end{table}

\paragraph{Implementation details.}
All models share the production Transformer backbone: 12 Transformer layers,
$d = 512$ hidden dimensions, 4 attention heads, maximum
sequence length $C = 1024$. We train with Adam optimizer and learning rate $10^{-3}$ with cosine annealing, batch
size 800. SIREN-RoPE uses a 2-layer SIREN branch and a 2-layer DNN branch (each 64 hidden units),
adding $\sim$0.2\% extra parameters over ordinal RoPE.

\paragraph{Evaluation metrics.}
We report Normalized Entropy (NE, lower is better) and AUC (higher is
better) for each of three engagement tasks, defined as follows:
\begin{itemize}
  \item \textbf{Contribution}: a binary label that equals 1 if any engagement
    action---Like, Share, Comment, Message, or Save---is observed on the post,
    and 0 otherwise. It captures the broadest notion of active engagement.
  \item \textbf{Like}: a binary label for the explicit Like action alone.
  \item \textbf{LongDwell}: a binary dwell-time engagement label derived from a
    post-type-specific threshold calibrated on historical data (e.g.,
    $\geq\!28$\,s for video posts, $\geq\!15$\,s for text posts). It captures
    passive but sustained attention.
\end{itemize}
NE measures calibration quality of the predicted engagement probability;
AUC measures ranking quality. Reporting both dimensions is essential because
a model can improve ranking (AUC) without improving calibration (NE), or
vice versa.

\subsection{Main Results}

Table~\ref{tab:main} reports held-out engagement prediction performance.
Because all models share the same backbone, features, and training procedure
(Table~\ref{tab:model_comparison}), every difference in the table is
attributable solely to how---or whether---the timestamp is used.

Two findings stand out before examining SIREN-RoPE.  First,
\textbf{Timestamp-as-Feature}---which appends the timestamp to item sequence
features while leaving the rotation angle purely ordinal---performs no better
than Ordinal RoPE on any metric, and is slightly worse on calibration (NE).
This mirrors the ablation finding (Section~\ref{sec:experiments}) that temporal signals routed through the embedding dimension fail to help and can interfere with content semantics.  Second,
\textbf{TO-RoPE}---which applies RoPE-style inverse frequencies to the
timestamp rather than ordinal position---only slightly improves over
Ordinal RoPE on LongDwell task, confirming that adding static time frequencies in a fixed
frequency schedule only helps on a very specific task.

\textbf{SIREN-RoPE}, by contrast, improves over Ordinal RoPE consistently
across all six metric--task combinations: it reduces Contribution NE by
0.0024 (0.6206$\to$0.6182), improves Like AUC by 0.0011
(0.9238$\to$0.9249), and achieves the largest gain on LongDwell AUC
($+$0.0036, 0.7597$\to$0.7633), consistent with the hypothesis that
engagement is temporally structured and therefore benefits from a rich rotation dimension.

\begin{table}[t]
  \caption{Engagement prediction on the production social feed dataset.
    NE: lower is better; AUC: higher is better. Best results are
    \textbf{bolded}; second-best is \underline{underlined}.}
  \label{tab:main}
  \centering
  \small
  \begin{tabular}{lcccccc}
    \toprule
    & \multicolumn{2}{c}{Contribution} & \multicolumn{2}{c}{Like}
    & \multicolumn{2}{c}{LongDwell} \\
    \cmidrule(r){2-3}\cmidrule(r){4-5}\cmidrule(r){6-7}
    Model & NE & AUC & NE & AUC & NE & AUC \\
    \midrule
    Ordinal RoPE (production) & \underline{0.6206} & \underline{0.9102} & \underline{0.5985} & \underline{0.9238} & 0.8362 & 0.7597 \\
    Timestamp-as-Feature      & 0.6218 & 0.9098 & 0.5997 & 0.9233 & 0.8350 & 0.7603 \\
    TO-RoPE                   & 0.6218 & 0.9095 & 0.5999 & 0.9231 & \underline{0.8349} & \underline{0.7613} \\
    \midrule
    SIREN-RoPE (ours)         & \textbf{0.6182} & \textbf{0.9115} & \textbf{0.5963} & \textbf{0.9249} & \textbf{0.8334} & \textbf{0.7633} \\
    \bottomrule
  \end{tabular}
\end{table}

\subsection{Ablation Study}

Table~\ref{tab:ablation} presents a systematic ablation across six variants,
grouped into three questions about the rotation dimension.

\begin{table}[t]
  \caption{Ablation study on the production social feed dataset across all
    six metric--task combinations. NE: lower is better; AUC: higher is
    better. Best per column in \textbf{bold}; second-best
    \underline{underlined}. Groups separated by horizontal rules.}
  \label{tab:ablation}
  \centering
  \small
  \begin{tabular}{lcccccc}
    \toprule
    & \multicolumn{2}{c}{Contribution} & \multicolumn{2}{c}{Like}
    & \multicolumn{2}{c}{LongDwell} \\
    \cmidrule(r){2-3}\cmidrule(r){4-5}\cmidrule(r){6-7}
    Variant & NE & AUC & NE & AUC & NE & AUC \\
    \midrule
    Semantic rotation (in/out-network) & 0.6192 & 0.9099 & 0.5972 & 0.9234 & 0.8356 & 0.7604 \\
    \midrule
    SIREN-RoPE + \texttt{time\_in\_year} only           & 0.6194 & \underline{0.9115} & 0.5973 & \underline{0.9250} & 0.8350 & 0.7629 \\
    DNN-only + \texttt{time\_in\_year} only            & 0.6191 & \textbf{0.9118}   & 0.5972 & \textbf{0.9252}   & 0.8346 & \underline{0.7630} \\
    SIREN-only + full temporal ($f_{\mathrm{DNN}}\!=\!0$)  & 0.6251 & 0.9095 & 0.6029 & 0.9232 & 0.8409 & 0.7600 \\
    DNN-only + full temporal ($f_{\mathrm{sin}}\!=\!0$)    & \textbf{0.6180} & 0.9111 & \textbf{0.5961} & 0.9244 & \textbf{0.8327} & 0.7627 \\
    Full SIREN-RoPE (ours)                              & \underline{0.6182} & \underline{0.9115} & \underline{0.5963} & 0.9249 & \underline{0.8334} & \textbf{0.7633} \\
    \bottomrule
  \end{tabular}
\end{table}

\paragraph{(1)~Cyclical features improve calibration.}
Replacing the scalar \texttt{time\_in\_year} with the full five-dimensional
$(\cos,\sin)$ decomposition consistently improves NE. The $(\cos,\sin)$
representation maps each cycle onto a unit circle, eliminating the
discontinuous jump (e.g., 86\,399\,s back to 0 at midnight) that a raw
scalar imposes on the network.

\paragraph{(2)~SIREN branch: useful for undiscovered periodicities.}
When cyclical features are already provided, DNN-only with full temporal
inputs achieves the best NE, since the $(\cos,\sin)$ pairs already supply
the periodic structure SIRENs are designed to learn. We retain the SIREN
branch not for the known daily/weekly cycles, but for its capacity to
\emph{autonomously discover hidden periodicities} absent from the manual
feature decomposition.

\paragraph{(3)~Semantic rotation: static signals belong in embeddings.}
Replacing the temporal input with a binary in/out-of-network feature
performs no better than ordinal RoPE. A static social-graph property does
not change across time or position, so the dynamic rotation dimension
offers no advantage over the embedding space for encoding it. The rotation
dimension is most productive with signals that are inherently
\emph{relational and dynamic}.

\paragraph{Efficiency analysis.}
SIREN-RoPE adds a negligible overhead in training ($\sim$1.4\% wall-clock
time increase over ordinal RoPE at batch size 800, sequence length 1024) and
inference ($\sim$1.8\%). The extra parameters constitute 0.2\% of total model
parameters. This confirms that the performance improvements come at minimal
computational cost.

\section{Discussion}
\label{sec:discussion}

\paragraph{Rotation as an expressive dimension.}
Our results suggest the rotation manifold is a structured expressive channel rather than a mere positional scaffold. 
Much like the imaginary axis in algebra, a learnable rotation dimension provides an orthogonal space to encode 
relationships invisible to the embedding space alone. Table~\ref{tab:ablation} confirms this: while injecting temporal 
features into embeddings is ineffective or detrimental, routing them through the rotation manifold yields consistent gains. 
This identifies rotation as a qualitatively distinct representation mode, suggesting a unified algebraic structure where 
semantic (embedding) and dynamic (rotation) dimensions provide non-redundant information.

\paragraph{Complementarity and Generalization.}
The stability of the learnable gate $\lambda$ demonstrates that ordinal recency and absolute temporal signals are 
complementary. This framework extends beyond recommendation to general sequence modeling; for instance, in NLP, 
the rotary angle could be conditioned on syntactic depth or part-of-speech tags. Such "content-aware" positional geometry 
allows the attention mechanism to modulate similarity based on structural or semantic context rather than simple proximity, 
offering a promising direction for more flexible transformer architectures.

\section{Conclusion and Future Work}
\label{sec:conclusion}

We presented \textbf{SIREN-RoPE}, which populates the RoPE rotation manifold
with temporal signals via a dual-branch SIREN--DNN network fused with ordinal
position through a learnable gate. The broader argument is that the rotation
space is an underexplored second dimension of attention expressivity---the
dynamic (imaginary) complement to the semantic (real) embedding space.
Experiments on a production social feed dataset confirm consistent gains
across calibration and ranking objectives at negligible computational cost,
and ablations confirm that the rotation dimension carries signal complementary
to and independent of the embedding space.

\paragraph{Future work.}
Open directions include: (1) a theoretical characterization of what function
classes the rotation manifold can express; (2) conditioning $f_\phi$ on
token-type or categorical features for semantic-temporal rotation and potential application in LLM;
(3) adapting the framework to cross-attention with separate angle functions
per side; and (4) unifying positional encoding across modalities via a shared learnable rotation space.

\subsection{Broader Impact Statement}

The proposed SIREN-RoPE framework introduces a more expressive method for encoding
temporal and semantic signals in Transformer architectures. While primarily a
fundamental improvement in representation learning, its applications---particularly
in large-scale recommendation and ranking systems---carry significant social
implications.

\paragraph{Positive impacts.}
By enabling models to more accurately capture complex periodicities and user behavior
patterns, SIREN-RoPE can lead to more relevant and efficient content discovery.
The unified framework may reduce the need for manual, feature-specific
engineering, potentially lowering the barrier to developing sophisticated sequential
models.

\paragraph{Potential negative impacts.}
As with any advancement in ranking and recommendation, improved model expressivity
could inadvertently reinforce filter bubbles or algorithmic biases if the underlying
semantic signals contain historical prejudices.

\subsection{Limitations}

\paragraph{Extrapolation constraints.}
The ``content-aware'' nature of SIREN-RoPE may be constrained by the distribution
of temporal and semantic features seen during training. Generalization to
out-of-distribution time intervals or entirely new semantic categories remains an
open question requiring further empirical study.

\begin{ack}
The authors thank colleagues on the production ranking team for helpful
discussions and infrastructure support. We also thank the anonymous reviewers
for their constructive feedback.
\end{ack}

\section*{References}

\bibliographystyle{abbrvnat}
\bibliography{base}


\appendix

\section{Attention Modulation of Ordinal RoPE}
\label{sec:rope_analysis}

\begin{figure}[h]
  \centering
  \includegraphics[width=0.85\columnwidth]{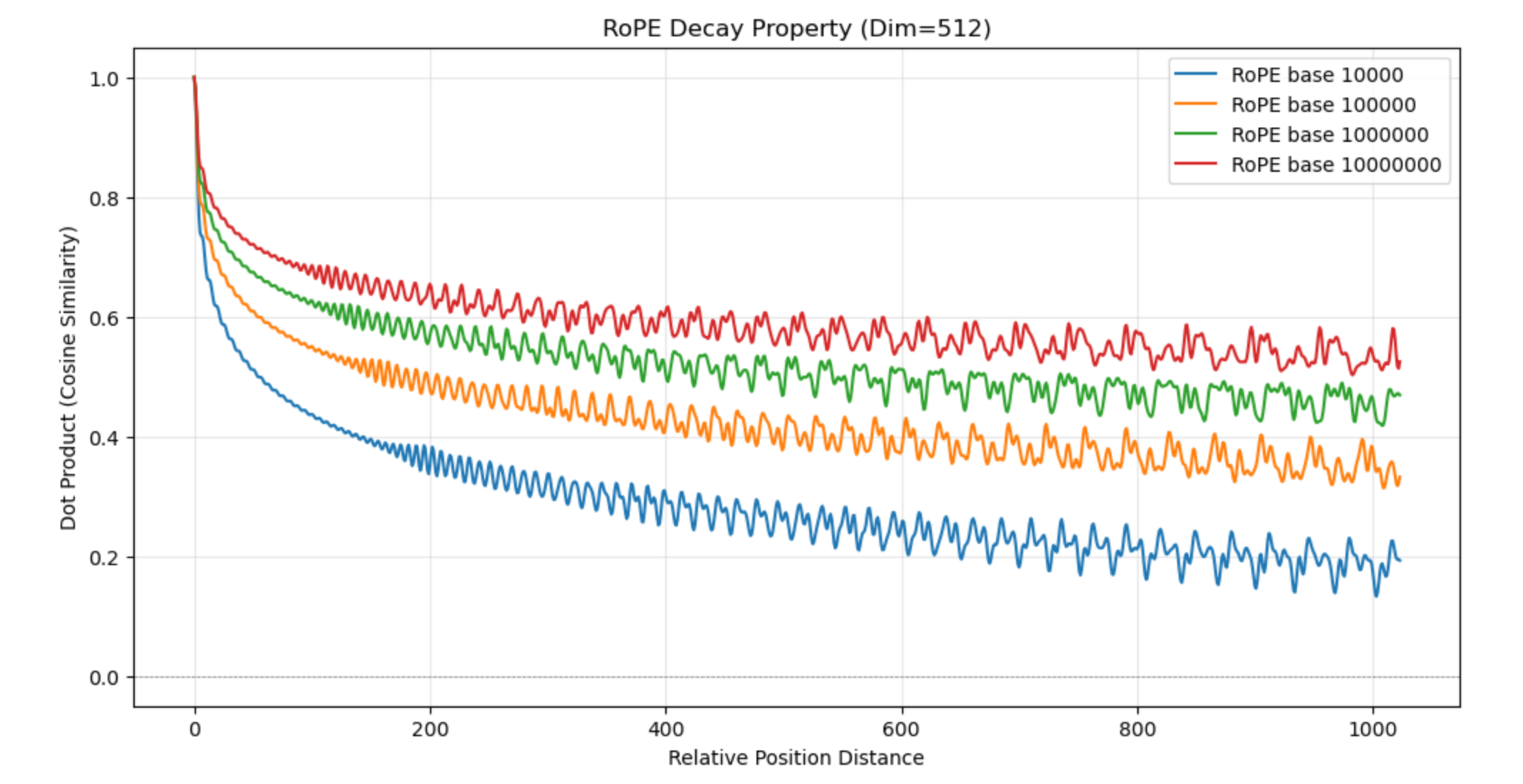}
  \caption{Attention score between a query at position 0 and keys at ordinal
    positions 0--1023, computed with a 512-dimensional unit vector
    ($\mathbf{q} = \mathbf{k} = \mathbf{1}/\sqrt{512}$) under four RoPE bases
    ($10^4$, $10^5$, $10^6$, $10^7$). All bases produce a monotone global decay
    with small high-frequency oscillations whose period grows with the base.}
  \label{fig:rope_analysis}
\end{figure}

Figure~\ref{fig:rope_analysis} visualizes what ordinal RoPE actually computes
in a recommendation context. We fix a 512-dimensional unit vector as both
query and key, place the query at position 0, and sweep the key over ordinal
positions 0--1023 for four commonly used bases.  Two properties are
immediately apparent.

\paragraph{Global recency decay.}
Regardless of base, the attention score decays monotonically with ordinal
distance. This is the primary inductive bias ordinal RoPE injects into
recommendation models: recent items receive higher attention weight, acting
as a soft recency prior.

\paragraph{High-frequency oscillations without semantic meaning.}
Superimposed on the decay are sinusoidal oscillations whose frequency is
determined by the base. In natural language, these oscillations encode
syntactic and semantic distance structure. In recommendation, however, ordinal
position has no intrinsic periodic meaning---items are indexed by interaction
order, not by any cyclical process---so the oscillation pattern carries no
useful signal. Instead it acts as structured noise injected into the attention
logits.

\paragraph{Ordinal RoPE as a regularizer.}
This structured noise turns out to be beneficial in a different way: it
stabilizes training. In our experiments, replacing no-RoPE attention with
ordinal RoPE allowed the learning rate to be increased from $10^{-4}$ to
$10^{-3}$ without training instability or loss divergence. We therefore
attribute two roles to ordinal RoPE in recommendation: (1)~a \emph{recency
prior} encoded in the global decay, and (2)~an \emph{implicit regularizer}
encoded in the oscillatory noise. SIREN-RoPE preserves both roles---through
the learnable ordinal gate $\lambda$---while additionally populating the
rotation manifold with genuine temporal semantics.

\paragraph{Base sensitivity in production.}
To understand whether the choice of base matters for recommendation quality,
we swept all four bases ($10^4$, $10^5$, $10^6$, $10^7$) on our production
model and found the downstream NE and AUC metrics to be nearly insensitive to
this choice. The differences across bases are within noise, consistent with
the observation in Figure~\ref{fig:rope_analysis} that all bases produce the
same qualitative decay structure. We selected $\mathrm{base} = 10^6$ for our
production deployment based on a negligible but consistent edge in held-out
metrics.

\section{SIREN-RoPE Temporal Attention Sweep}
\label{sec:time_sweep}

\begin{figure}[h]
  \centering
  \includegraphics[width=\columnwidth]{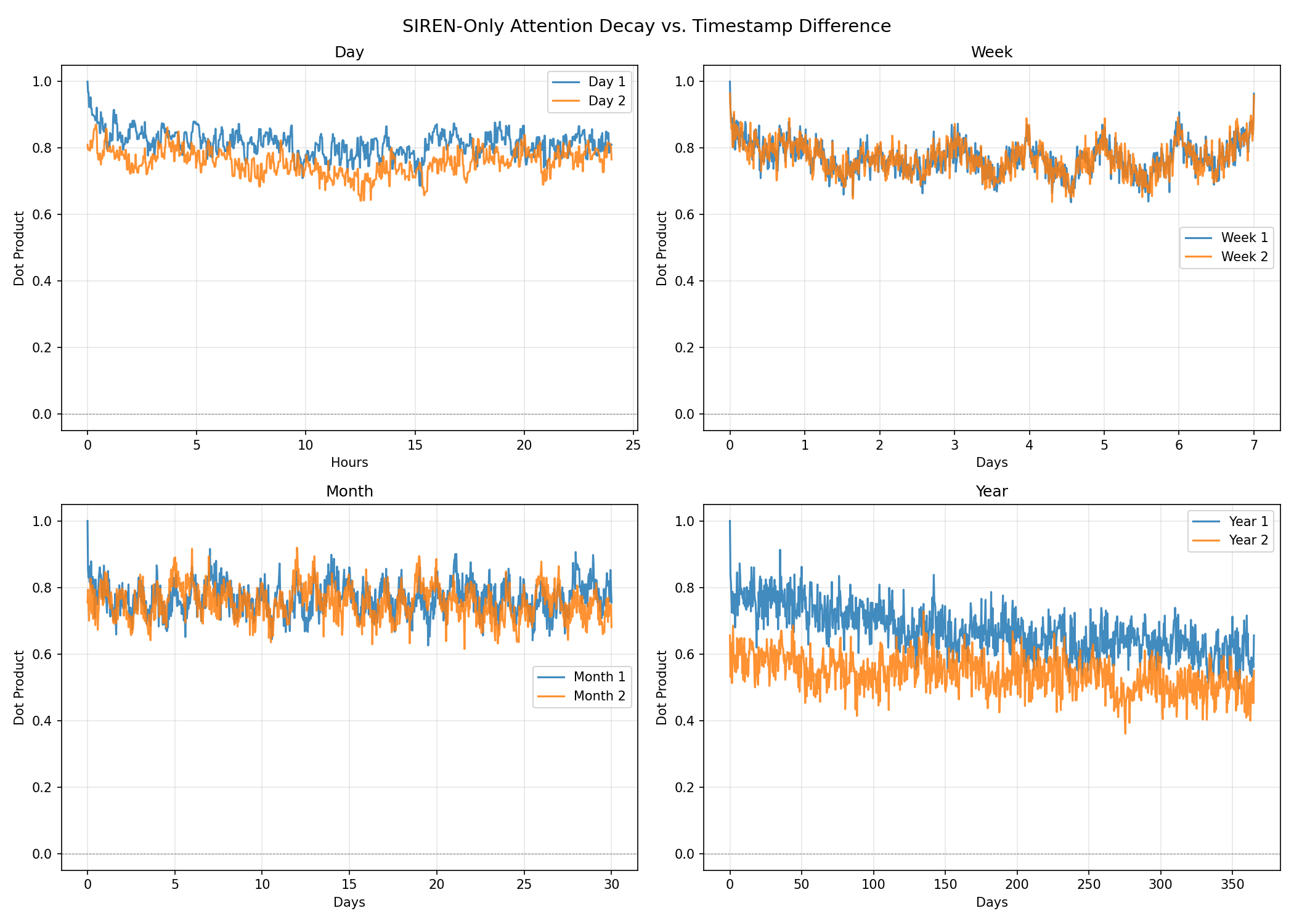}
  \caption{Attention score of a SIREN-RoPE module (weights extracted from the
    production model) as the key timestamp is swept across four time spans,
    using a 128-dimensional unit vector for both query and key. Each panel
    overlaps two consecutive periods (day 1 vs.\ day 2; week 1 vs.\ week 2;
    month 1 vs.\ month 2; year 1 vs.\ year 2) to reveal repeating structure.}
  \label{fig:sweep_time}
\end{figure}

To visualize what temporal structure SIREN-RoPE has actually learned, we
extract the trained weights from the production SIREN-RoPE module and build a
standalone inference unit that accepts a query timestamp, a key timestamp, and
128-dimensional query/key unit vectors, and returns the rotary-modulated
attention score. We fix the query at a reference point and sweep the key
timestamp across four spans---one day, one week, one month, and one year---
overlapping two consecutive periods in each panel of
Figure~\ref{fig:sweep_time} to expose repeating structure.

\paragraph{Strong intra-day periodicity.}
The day-over-day panel shows that the attention curves for day~1 and day~2
align closely, confirming that the SIREN branch has autonomously recovered
a sub-daily (hourly) oscillation pattern that repeats every 24 hours---directly from the training signal, without any explicit hour-of-day feature.

\paragraph{Strong weekly periodicity.}
The week-over-week panel shows near-perfect overlap of attention scores between week~1 and week~2.
The model captures the weekday/weekend behavioral split as a recurring
weekly rhythm encoded in the rotation angle. This weekly structure is the most dominant periodicity the network discovers, consistent with pronounced weekday/weekend engagement patterns in social feed data.

\paragraph{No monthly (30 days) periodicity.}
The month-over-month panel shows no alignment between month~1 and month~2:
the curves diverge immediately and do not repeat. This is expected---the dominant weekly period does not divide evenly into 30 days, so no recurring monthly pattern emerges.

\paragraph{Yearly decay only.}
The year-over-year panel shows a smooth, monotone decay in attention score as
the key recedes further into the past, with no recurring annual pattern.
This long-range behavior is captured by the aperiodic DNN branch rather than
the SIREN branch, confirming the intended division of labor: SIREN encodes
high-frequency periodicities (daily, weekly) while the DNN captures low-frequency
monotone trends (recency decay).

\begin{figure}[h]
  \centering
  \includegraphics[width=0.9\columnwidth]{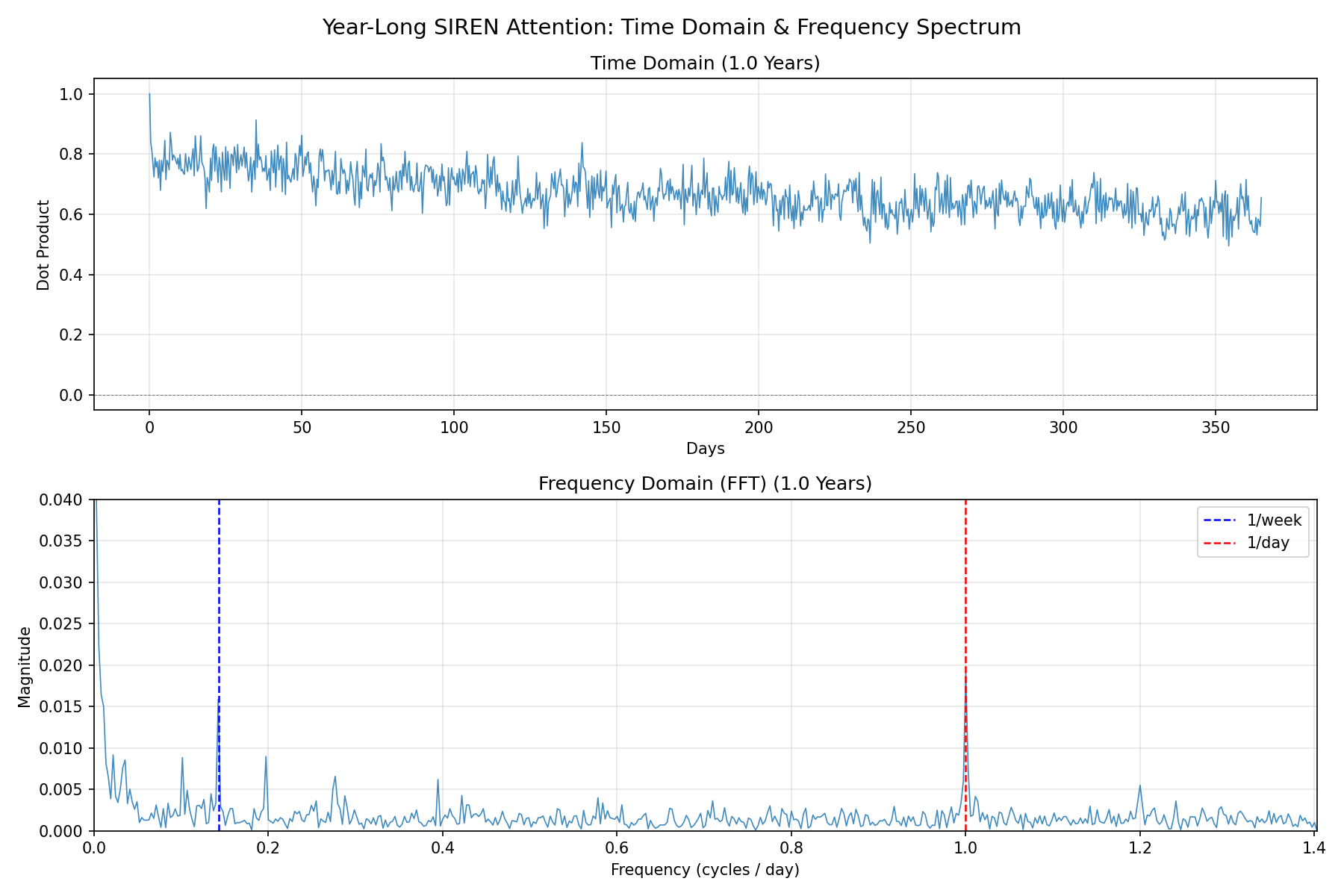}
  \caption{Year-long SIREN-RoPE attention in the time domain (top) and its
    FFT magnitude spectrum (bottom). Vertical dashed lines mark the weekly
    ($\frac{1}{7}$\,cycles/day) and daily (1\,cycle/day) frequencies. Both
    frequencies exhibit prominent peaks; sub-harmonics at $\frac{1}{2}$,
    $\frac{1}{3}$, \ldots cycles/day are also visible.}
  \label{fig:year_fft}
\end{figure}

\paragraph{Frequency-domain analysis.}
To quantify the learned periodicities, we apply the Fast Fourier Transform
(FFT) to the year-long attention curve; Figure~\ref{fig:year_fft} shows the
result.

The FFT spectrum reveals two dominant peaks: one at $\approx 0.143$\,cycles/day
($= 1/7$\,day$^{-1}$, the weekly frequency) and one at exactly
1\,cycle/day (the daily frequency).  Both are recovered without any
explicit period specification---the model learned these rhythms purely from the
training signal. These are consistent with common social-feed behavioral rhythms such as
weekend-only posting or mid-week engagement surges, and are naturally captured
by the harmonic richness of the SIREN branch.

\begin{figure}[h]
  \centering
  \includegraphics[width=\columnwidth]{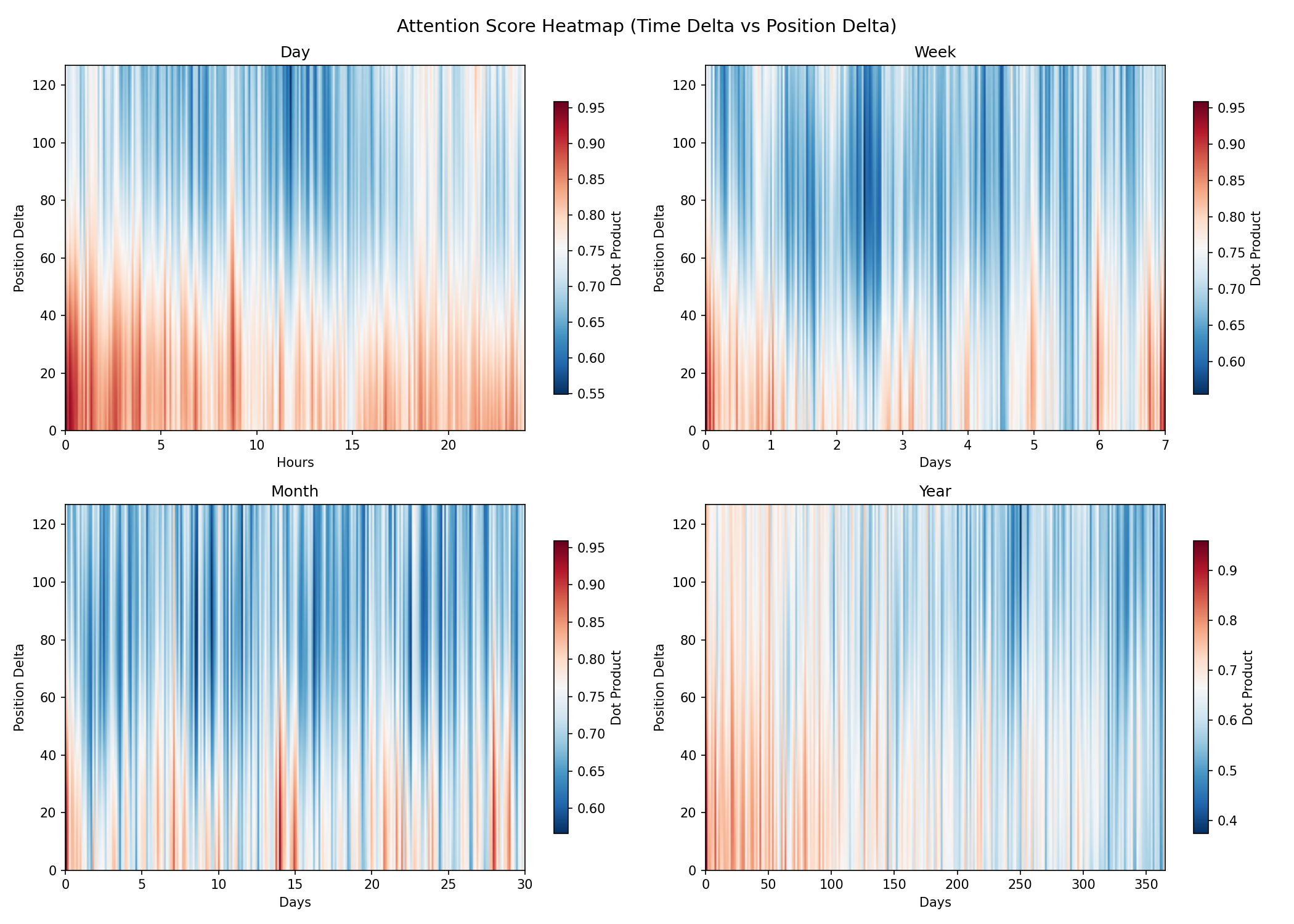}
  \caption{2D attention score heatmap as a function of key ordinal position
    (y-axis, 0--120) and key timestamp (x-axis) for four time spans.
    Color encodes attention magnitude (brighter = higher).
    The ordinal axis exhibits monotone decay irrespective of the time span;
    the temporal axis exhibits clear intra-day and intra-week oscillations
    in the short-span panels.}
  \label{fig:attn_heatmap}
\end{figure}

\paragraph{Joint ordinal--temporal attention landscape.}
Figure~\ref{fig:attn_heatmap} provides a unified view by sweeping \emph{both}
dimensions simultaneously: the ordinal position of the key (y-axis, 0--120)
and the key timestamp (x-axis) across four time spans (1 day, 1 week, 30
days, 365 days).

Two orthogonal patterns are immediately apparent in the heatmap.

\textbf{Ordinal axis: pure monotone decay.}
Across all four panels, attention decreases smoothly from ordinal position 0
(most recent) to 120 (oldest) with no oscillatory structure. This confirms
that the ordinal component $p_i \cdot \theta_j \cdot \lambda$ contributes a
clean recency prior and nothing else---its role is solely to weight recent
items more heavily, not to encode any periodic information.

\textbf{Temporal axis: oscillation reflecting human behavior.}
In the 1-day and 1-week panels the heatmap exhibits vivid vertical banding:
bright and dark stripes repeat with a period of roughly 24 hours (intra-day
engagement cycles) and 7 days (weekday/weekend rhythm). The oscillation is
consistent across all ordinal positions, confirming it originates entirely
from the temporal (SIREN) component $f_\phi(T_i)_j \cdot \omega^s_j$ rather
than from the ordinal component. In the 30-day panel the daily banding remains
visible but the weekly banding broadens; in the 365-day panel both fine-grained
oscillations fade out and the global temporal decay from the DNN branch
becomes visible.

Taken together, the time-domain sweeps, frequency-domain analysis, and joint
heatmap confirm that SIREN-RoPE learns a multi-resolution temporal geometry
directly in the rotation manifold---sharp daily and weekly oscillations
reflecting real user behavioral cycles, harmonic sub-peaks at integer-day
periods, and a smooth recency decay at the scale of months and
years---all without manual specification of these time scales except the feature engineering.


\newpage

\end{document}